\documentclass[letterpaper,journal]{IEEEtran}
\usepackage{amsmath,amsfonts}
\usepackage{algorithm}
\usepackage{algorithmic}
\usepackage{array}
\usepackage[caption=false,font=normalsize,labelfont=sf,textfont=sf]{subfig}
\usepackage{textcomp}
\usepackage{stfloats}
\usepackage{url}
\usepackage{verbatim}
\usepackage{graphicx}
\usepackage{cite}
\usepackage{hyperref}
\usepackage{orcidlink}
\usepackage{booktabs}
\usepackage{multirow}
\usepackage[table]{xcolor}

\begin{document}

\title{TFM Dataset: A Novel Multi-task Dataset and Integrated Pipeline for Automated Tear Film Break-Up Segmentation}
\author{Guangrong Wan\textsuperscript{\orcidlink{0009-0003-6891-3463}}, Jun Liu\textsuperscript{\orcidlink{0000-0001-7390-8958}} \IEEEmembership{Member, IEEE},
Qiyang Zhou\textsuperscript{\orcidlink{0009-0006-2713-0085}},
Tang Tang\textsuperscript{\orcidlink{0009-0005-6657-6429}}, \\
Lianghao Shi\textsuperscript{\orcidlink{0009-0000-6597-0700}}, 
Wenjun Luo\textsuperscript{\orcidlink{0009-0007-8365-5713}},
TingTing Xu\textsuperscript{\orcidlink{0000-0002-4468-1413}}
\thanks{This work was supported by the Chongqing Research Program of Basic Research and Frontier Technology under Grant No. CSTB2025NSCQ-GPX1309 and CSTB2024NSCQ-MSX0805.(Corresponding author: Jun Liu.)}
\thanks{Guangrong Wan, Qiyang Zhou, Tang Tang, Lianghao Shi, and Wenjun Luo are with the School of Computer Science and Technology (National Exemplary Software School), Chongqing University of Posts and Telecommunications, Chongqing 400065, China (e-mails: 2021214021@stu.cqupt.edu.cn; 2023214550@stu.cqupt.edu.cn; 2022214265@stu.cqupt.edu.cn; 2022214069@stu.cqupt.edu.cn; 2022214293@stu.cqupt.edu.cn).}
\thanks{Jun Liu and Tingting Xu are with the School of Computer Science and Technology (National Exemplary Software School), Chongqing University of Posts and Telecommunications, Chongqing 400065, China (e-mails: junliu@cqupt.edu.cn; xutt@cqupt.edu.cn).}
}
\maketitle

\begin{abstract}
Tear film break-up (TFBU) analysis is critical for diagnosing dry eye syndrome, but automated TFBU segmentation remains challenging due to the lack of annotated datasets and integrated solutions. This paper introduces the Tear Film Multi-task (TFM) Dataset, the first comprehensive dataset for multi-task tear film analysis, comprising 15 high-resolution videos (totaling 6,247 frames) annotated with three vision tasks: frame-level classification ('clear', 'closed', 'broken', 'blur'), Placido Ring detection, and pixel-wise TFBU area segmentation. 
Leveraging this dataset, we first propose TF-Net, a novel and efficient baseline segmentation model. TF-Net incorporates a MobileOne-mini backbone with re-parameterization techniques and an enhanced feature pyramid network to achieve a favorable balance between accuracy and computational efficiency for real-time clinical applications. We further establish benchmark performance on the TFM segmentation subset by comparing TF-Net against several state-of-the-art medical image segmentation models.
Furthermore, we design TF-Collab, a novel integrated real-time pipeline that synergistically leverages models trained on all three tasks of the TFM dataset. By sequentially orchestrating frame classification for BUT determination, pupil region localization for input standardization, and TFBU segmentation, TF-Collab fully automates the analysis. Experimental results demonstrate the effectiveness of the proposed TF-Net and TF-Collab, providing a foundation for future research in ocular surface diagnostics. Our code and the TFM datasets are available at \href{https://github.com/glory-wan/TF-Net}{\textit{https://github.com/glory-wan/TF-Net}}
\end{abstract}


\section{Introduction}
\label{sec:introduction}
Dry eye syndrome (DES) is one of the most prevalent ophthalmic conditions globally, significantly impacting patients' quality of life by causing discomfort, visual disturbance, and potential damage to the ocular surface\cite{bg0, bg1, survey}. A cornerstone of DES diagnosis, particularly for evaporative dry eye, is the assessment of tear film stability\cite{bg2}, quantitatively measured by the Tear Film Break-Up Time (BUT). The Fluorescein Break-Up Time (FBUT) test is the clinical gold standard\cite{fluorescein}. It involves instilling fluorescein dye and observing the tear film under a blue light source. The clinician records the time between a complete blink and the first appearance of a dark spot (indicating tear film break-up), known as the BUT. However, the conventional approach to tear film break-up (TFBU) analysis relies heavily on manual observation and annotation by clinical experts. This process is not only time-consuming and labor-intensive but also inherently subjective, leading to considerable inter- and intra-observer variability\cite{placido}. The pursuit of automation and objectivity has thus become a major focus in ocular surface diagnostics.

The advancement of computer vision and deep learning offers a promising pathway towards automating and objectifying TFBU analysis. Automated systems have the potential to provide consistent, quantitative measurements of break-up areas\cite{tra1, Non-invasive}, leading to more reliable and reproducible BUT calculations and detailed mapping of tear film dynamics. Despite this potential, the development of robust AI-driven solutions for this task has been severely hampered by a critical bottleneck: \textbf{the lack of a large-scale, publicly available dataset with high-quality, multi-task annotations for the stability analysis of tear film.} Most existing studies rely on private, limited datasets, often annotated for a single purpose\cite{conv8, deep9, con15} (e.g., classification only). This absence prevents fair benchmarking of algorithms, hinders the development of integrated diagnostic pipelines, and ultimately slows down progress in the field.

\begin{figure*}[!t]
    \centering
    \includegraphics[width=1\textwidth]{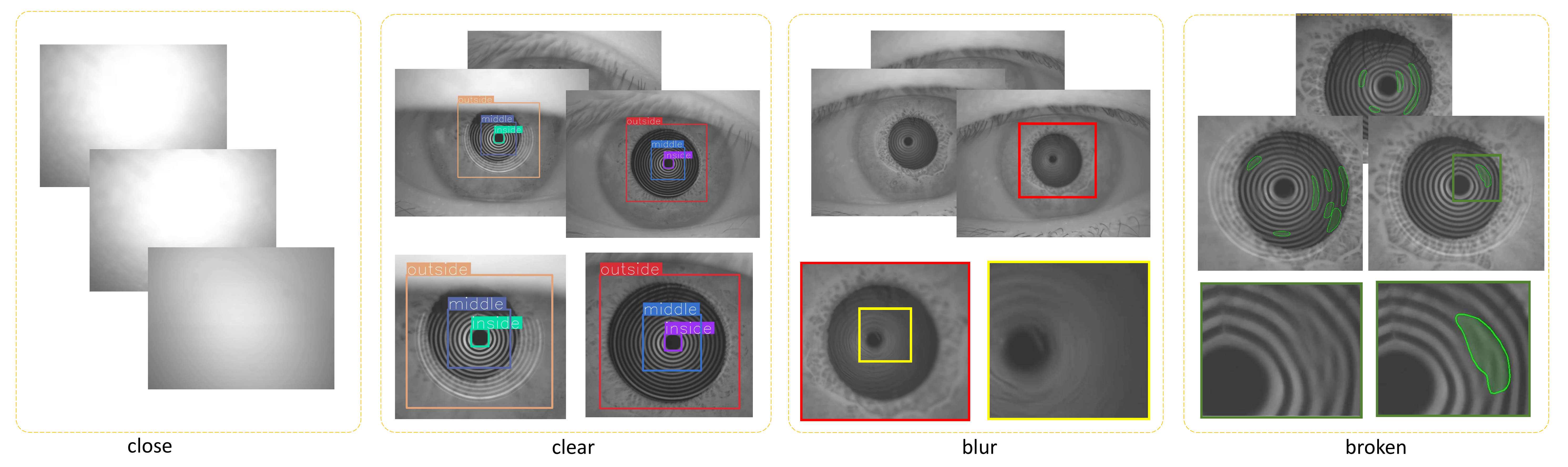}
    \caption{
        Overview of the Tear Film Multi-task (TFM) Dataset composition, illustrating the distribution and relationships between the three annotation tasks: classification (TF-Cls), object detection (TF-Det), and segmentation (TF-Seg).
    }
    \label{fig:tf_dataset}
\end{figure*}

\begin{table*}
  \centering
  \begin{tabular}{ccccc}
    \toprule
     Label Type     & Category   & Number    & Resolution    & Ratio(train:val:test)  \\
    \midrule
     TF-Cls, Classification 
                    & 4     & 6,247     & 3632 * 2760   & 4,687:561:999    \\
     TF-OD, Object Detection(Placido Ring)   
                    & 3     & 4,736     & 3632 * 2760   & 3,546:430:760 \\
     TF-Seg, Segmentation(Tear film Broken Area)       
                    & 1     & 873       & 3632 * 2760/1498 x 1337(TF-crop)   & 678:69:126 \\

    \bottomrule
  \end{tabular}
  \caption{
        Statistics of the Tear Film Multi-task (TFM) Dataset.
  }
  \label{tab:tf-dataset}
\end{table*}

Furthermore, existing automated approaches often address TFBU analysis in isolation. A simplistic segmentation-only model applied to raw clinical images faces several practical challenges: (1) \textbf{Irrelevant frames}: Videos contain frames that are unusable for analysis (e.g., during blinking, or motion-blurred), which must be filtered out. (2) \textbf{Region of Interest (ROI) localization}: The area of interest for precise TFBU quantification is within the placido ring reflection, which must be accurately detected and cropped to normalize the input for segmentation. (3) \textbf{Spatial mapping}: Clinical interpretation requires mapping the segmented break-up areas from the cropped ROI back to standardized coordinates for size and location analysis. An effective solution requires an integrated pipeline that can synergistically perform multiple tasks, such as filtering uninformative frames, standardizing the region of interest, and performing precise segmentation.

To bridge these critical gaps, we present three key contributions in this paper:

1) \textbf{We introduce the Tear Film Multi-task (TFM) Dataset}, the first comprehensive and public dataset of its kind. The TFM Dataset includes 6,247 high-resolution images extracted from 15 videos, each meticulously annotated for three distinct but complementary vision tasks: image classification (categorizing frames as 'clear', 'closed', 'broken', 'blur'), object detection (localizing the placido rings and pupil region), and pixel-wise semantic segmentation (delineating TFBU areas). This multi-task annotation scheme provides a rich resource for developing and evaluating holistic tear film analysis algorithms.

2) \textbf{We propose TF-Net, a strong baseline model for TFBU segmentation}. Leveraging the segmentation annotations of the TFM Dataset, TF-Net is designed with clinical deployment in mind, employing a MobileOne-mini backbone with reparameterization techniques and an enhanced feature pyramid network to achieve high accuracy while maintaining computational efficiency suitable for real-time processing.

3) \textbf{We design TF-Collab, a novel integrated pipeline for automated TFBU analysis}. TF-Collab orchestrates three specialized models (classification, detection, segmentation) in a synergistic, real-time sequence, demonstrating the practical utility of a multi-task approach. The pipeline seamlessly combines models trained for classification, detection, and segmentation to fully automate the TFBU analysis workflow: from frame selection and BUT calculation, through pupil region cropping, to final break-up segmentation and severity mapping back to standardized coordinates.

The remainder of this paper is organized as follows. Section 2 reviews related work on tear film analysis and medical image segmentation. Section 3 provides a detailed description of the TFM Dataset. Section 4 elaborates on the architecture of the TF-Net model and the design of the TF-Collab pipeline. Section 5 presents the experimental setup, results, and ablation studies. Finally, Section 6 discusses the implications of our work, acknowledges its limitations, and suggests directions for future research.

\section{Related Work}
The automated analysis of tear film break-up sits at the intersection of ophthalmology and computer vision. Accordingly, our review of related work encompasses both traditional clinical methods and learning-based approaches for ocular image analysis.
\begin{figure*}[!t]
    \centering
    \includegraphics[width=1\textwidth]{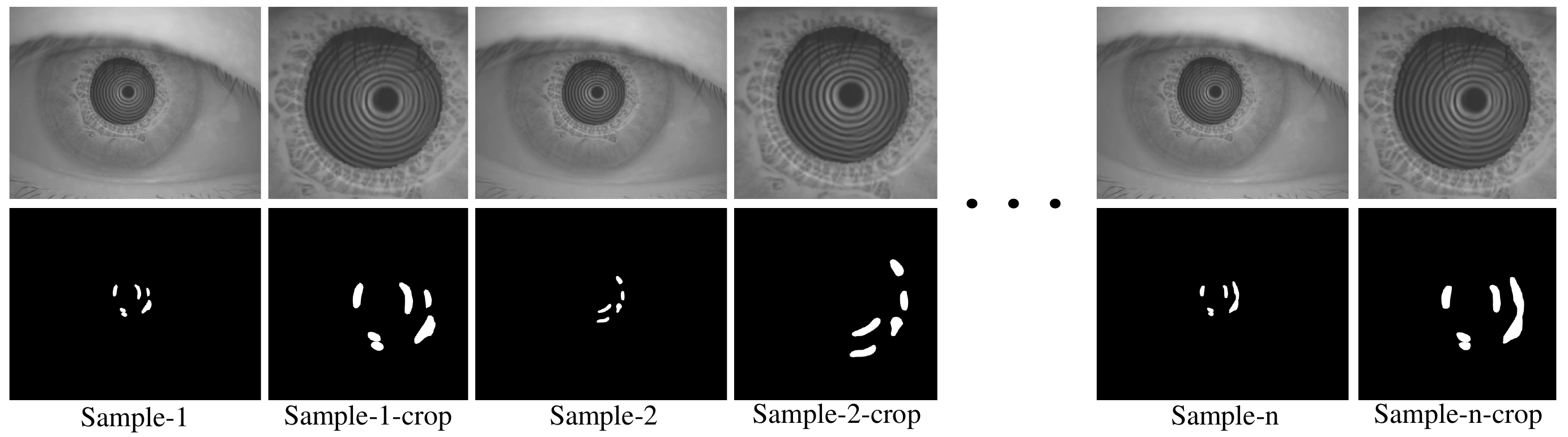}
    \caption{
        Sample visualization of the TF-Crop dataset from cropping strategy.. The first row displays the original full-resolution images (left) and their corresponding cropped versions (right), which are generated based on the "Outside" bounding boxes from the TF-Det dataset. The second row presents the visualizations of the pixel-wise TFBU segmentation masks for the respective images above.
    }
    \label{fig:crop-set}
\end{figure*}

\subsection{Deep Learning for Tear Film Analysis}
Deep learning technologies, particularly convolutional neural networks (CNNs), have gained widespread use in the measurement of tear film break-up time (TBUT)\cite{conv8 ,deep9, bg1, Non-invasive14, con15}. In the study by \cite{conv8}, the researchers adjusted the hyperparameters of GoogLeNet to extract image features from tear film video frames and performed four-class classification (normal, rupture, blinking, and noise). The BUT was determined by calculating the time interval between blinking and tear film rupture. In \cite{placido13}, the researchers added 3D convolutional layers to the CNN model to capture dynamic features of image sequences over time. Meanwhile, \cite{Non-invasive14} used a pretrained ResNet50 model, fine-tuned on tear film interference video data, and divided the tear film images into several cells, predicting the rupture location for each cell. In \cite{deep9}, a CNN model was proposed to identify the first tear film rupture frame in TBUT videos. However, these methods still have some limitations. For example, the models in \cite{conv8, deep9, con15} are only suitable for classification tasks. The addition of 3D convolutional layers in \cite{placido13} only improved classification accuracy. In clinical practice, it is important not only to identify the rupture time but also to measure the location and area of the tear film rupture, as this information provides valuable insights for diagnosis. Moreover, several of these studies \cite{conv8, deep9} relied on fluorescein-stained images or videos, which are inherently limited by the drawbacks of BUT. These methods, while reducing the need for manual observation and recording, still inherit the limitations of BUT, such as patient discomfort and interference with the tear film's natural state. To address these issues, we propose a non-invasive method using placido ring images, which offer higher contrast and clarity. Our method, TF-Collab, efficiently performs tear film frame classification, placido rings detection, and TFBU segmentation, providing comprehensive and systematic analysis of tear film stability.

\subsection{Medical Image Segmentation}
Medical image segmentation, which aims to assign pixel-wise labels to anatomical or pathological regions, is vital for computer-aided diagnosis. The field has been largely shaped by the evolution of convolutional neural networks (CNNs). The U-Net architecture\cite{unet}, with its symmetric encoder-decoder structure and skip connections, remains a foundational benchmark for biomedical segmentation. Subsequent variants like U-Net++\cite{unet++, unet++tmi} and Attention U-Net\cite{unetattention} further improved performance through nested connections and attention mechanisms.
More recently, the field has witnessed two significant trends. The first is the adoption of architectures inspired by natural language processing. Vision Transformers (ViTs)\cite{vit} have been successfully integrated, leading to hybrid models like TransUNet\cite{transunet} and Swin-Unet\cite{swin}, which capture long-range dependencies to achieve state-of-the-art results on many benchmarks. The second, and perhaps most prominent, trend is the rise of foundation models. The SAM\cite{SAM} and MedSAM\cite{MedSAM, MedSAM2} introduced a promptable, general-purpose segmentation model with zero-shot capabilities. However, their performance on specialized medical imaging tasks without fine-tuning is often suboptimal, prompting research into adapting it for medical domains. Simultaneously, self-supervised learning frameworks like DINO\cite{dinov3} have produced powerful visual features that serve as excellent backbones for downstream tasks, including segmentation, by learning robust representations from vast amounts of unlabeled data.
While these advanced models push the boundaries of accuracy, their high computational cost can hinder real-time clinical deployment. This has spurred parallel research into efficient and lightweight architectures. Techniques like neural architecture search and structural reparameterization, as exemplified by MobileOne\cite{mobileone}, are crucial for creating models that balance performance with speed.

\section{Tear Film Multi-task Dataset}
We acquired multiple high-resolution tear film videos captured using a Placido ring-based dry eye instrument. A total of 6,247 valid frames were extracted from these videos. Each frame was annotated for three high-level vision tasks: image classification ('Clear', 'Closed', 'Broken', and 'Blur'), object detection (for Placido rings and pupil region), and semantic segmentation (for tear film break-up area). Specifically, classification labels were assigned to all images. For all non-Closed frames (i.e., those displaying the Placido ring), bounding boxes were annotated to localize the Placido ring. Furthermore, all images labeled as "Broken" underwent pixel-level segmentation annotation for the TFBU area. The dataset composition and annotations are shown in Fig. \ref{fig:tf_dataset}. All annotations are provided using X-AnyLabeling\cite{X-AnyLabeling}.

\subsection{Dataset Collection and Annotation}
The dataset was constructed from 15 placido ring imaging videos. Through frame-by-frame extraction, we obtained 6,247 images, which were then randomly split into training (4,687 images), validation (561 images), and test (999 images) sets. All three subsets were consistently annotated for the three tasks mentioned above. Detailed statistics are provided in Table \ref{tab:tf-dataset}. The annotation criteria are as follows:

\textbf{Image Classification (TF-Cls).} According to the definition of Break-Up Time (BUT) -— the time interval between a blink and the first appearance of tear film break-up -— it is necessary to identify the moment of blinking, i.e., images classified as "Closed". Additionally, during actual clinical diagnosis, unavoidable blurring occurs due to patient eye tremors or blinking; these image frames are regarded as noise and classified as "Blur". The remaining clear open-eye images are further categorized into "Clear" and "Broken" based on the presence or absence of tear film break-up. Following these criteria, all 6,247 images were classified into four categories: 'Clear', 'Closed', 'Broken' and 'Blur'.

\textbf{Placido Rings Detection (TF-Det).} We annotated three types of placido ring on all non-closed category images(i.e., where the placido rings are visible): inside, middle, and outside. The "inside" annotation was defined by fitting the bounding box tangent to the innermost edge of the smallest ring. The "middle" annotation was defined by fitting the bounding box tangent to the outer edge of the fifth ring from the inside. The "outside" annotation encompassed the entire placido rings structure.

\textbf{Tear Film Break-Up Area Segmentation (TF-Seg).} Based on the imaging principles of the placido ring, tear film break-up is identified by the presence of distortions, discontinuities, or other abnormalities in the ring. This task is highly challenging due to a severe class imbalance; in the original images, background pixels constitute 99.82\% of the area, while TFBU regions account for only 0.18\%. Given this imbalance and the high resolution of the original images, we cropped both the images and their corresponding segmentation masks using the "Outside" detection bounding boxes. This preprocessing step created a focused region-of-interest (ROI) subset, termed \textbf{TF-Crop}. The cropping strategy effectively addresses the imbalance, increasing the proportion of TFBU pixels to 0.93\% (while background is reduced to 99.07\%), and standardizes the input for models. As shown in Fig. \ref{fig:crop-set} and demonstrated in our experiments, this strategy significantly enhances segmentation performance.

\section{METHODS}
\subsection{Problem Definition and TF-Collab Pipeline}
\textbf{Pipeline System Problem.}
The ultimate goal of our automated diagnostic system is to process a tear film video sequence $\mathcal{V} = \{I_1, I_2, ..., I_T\}$, where $I_t$ denotes the $t$-th frame, and to produce two key clinical outputs: 1) the Break-Up Time (BUT), and 2) a sequence of segmentation maps localizing the Tear Film Break-Up (TFBU) areas in their original spatial context. Formally, we define this as learning a function $\mathcal{F}_{sys}$ that maps the input video $\mathcal{V}$ to the outputs:
\begin{equation}
\mathcal{F}_{sys}(\mathcal{V}) = (t_{BUT}, \{\hat{S}_t\}_{t=1}^{T})
\end{equation}
where $t_{BUT} \in [1, T]$ is the frame index identified as the first break-up event, and $\hat{S}_t$ is the predicted TFBU segmentation mask for frame $I_t$, which is null for frames where no break-up is detected or frames are invalid (e.g., blurry). The challenge lies in the fact that $\mathcal{F}_{sys}$ must intelligently integrate the solutions to three interdependent sub-problems: frame quality assessment, region-of-interest localization, and precise pathological area segmentation.

To solve $\mathcal{F}_{sys}$, we propose \textbf{TF-Collab}, an integrated pipeline that synergistically orchestrates three specialized models, each trained for a sub-task defined on the TFM dataset. The pipeline, summarized in Algorithm 1, proceeds as follows for each frame $I_t$:

\begin{algorithm}[t]
\caption{TF-Collab Pipeline}
\label{alg:tf_collab}
\begin{algorithmic}[1]
\REQUIRE A video stream $\mathcal{V} = \{I_1, I_2, ..., I_T\}$
\ENSURE Break-Up Time $t_{BUT}$, Sequence of segmentation maps $\{\hat{S}_t\}$
\STATE Initialize $t_{BUT} \gets \emptyset$, $t_{onset} \gets 1$
\FOR{$t = 1$ \textbf{to} $T$}
    \STATE $c_t \gets f_{cls}(I_t)$
    \IF{$c_t = \text{`Closed'}$}
        \STATE $t_{onset} \gets t$
    \ELSIF{$c_t = \text{`Broken'}$ \textbf{and} $t_{BUT} = \emptyset$}
        \STATE $t_{BUT} \gets t - t_{onset}$
    \ENDIF
    \IF{$c_t \in \{\text{`Clear'}, \text{`Blur'}, \text{`Broken'}\}$}
        \STATE $B_t^{outside} \gets f_{det}(I_t)$
        \STATE $I_t^{crop} \gets \text{Crop}(I_t, B_t^{outside})$
        \STATE $\hat{M}_t \gets f_{seg}(I_t^{crop})$
        \STATE $\hat{S}_t \gets \text{MapBack}(\hat{M}_t, B_t^{outside})$
    \ELSE
        \STATE $\hat{S}_t \gets \emptyset$
    \ENDIF
\ENDFOR
\end{algorithmic}
\end{algorithm}

Algorithm 1 outlines the TF-Collab pipeline for automated TFBU analysis. The pipeline processes an input video stream $\mathcal{V}$ to compute the Break-Up Time ($t_{BUT}$) and generate a sequence of segmentation maps ($\{\hat{S}_t\}$). For each frame $I_t$, the classification model $f_{cls}$ first assigns a category label $c_t$. The BUT calculation module tracks the onset time $t_{onset}$ after a blink event (indicated by a 'Closed' frame) and computes $t_{BUT}$ upon the first occurrence of a 'Broken' frame.

\begin{figure*}[!t]
    \centering
    \includegraphics[width=1\textwidth]{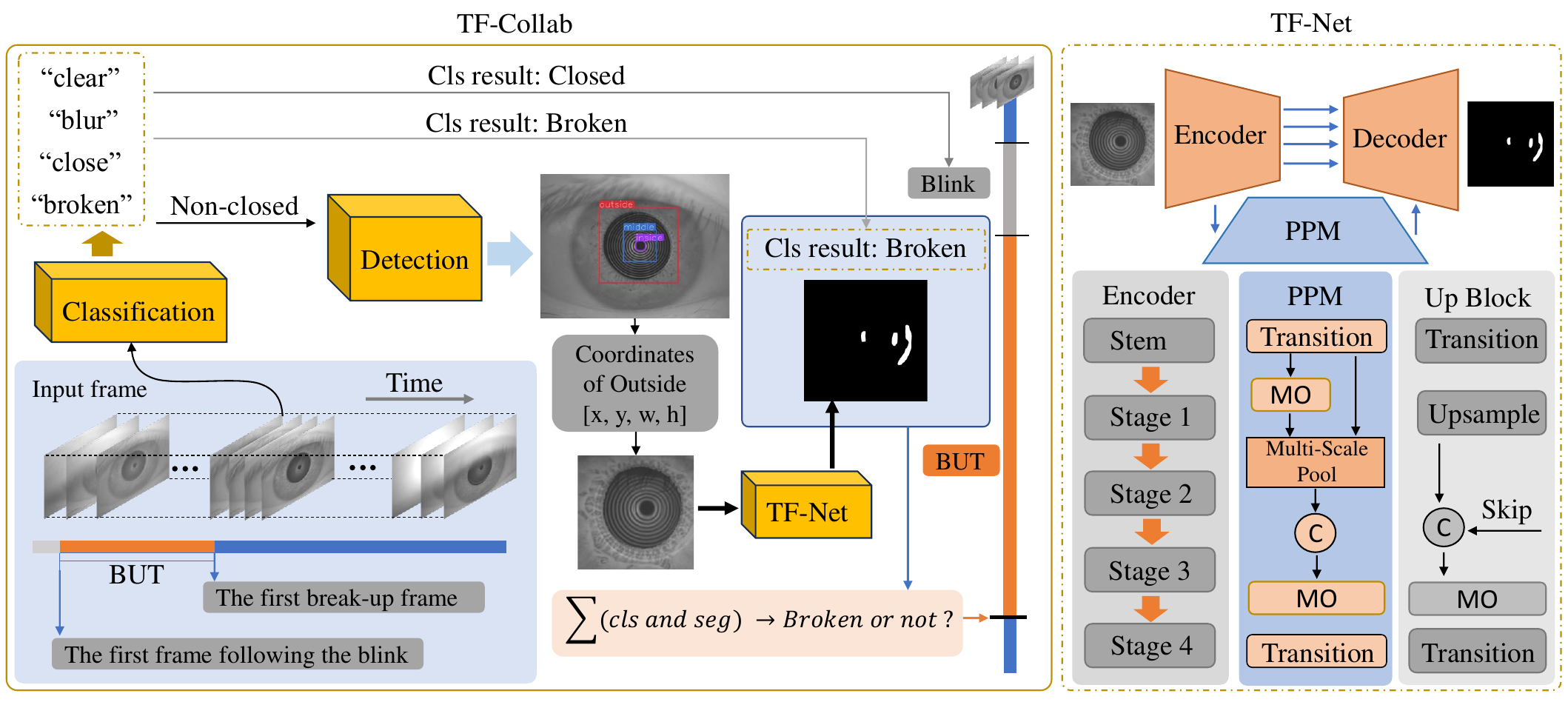}
    \caption{
         (a) The overall workflow of the proposed TF-Collab pipeline, which sequentially integrates frame classification, placido rings detection, and TFBU segmentation. (b) The detailed architecture of the proposed TF-Net model, featuring a MobileOne-mini encoder with re-parameterization, a Pyramid Pooling Module (PPM) for multi-scale context, and a decoder with skip connections for boundary refinement. 'MO' means mobileone block\cite{mobileone}.
    }
    \label{fig:tf_dino}
\end{figure*}

Frames classified as non-closed are subsequently processed by the placido rings detection model $f_{det}$, which localizes the region of interest by predicting the pupil box $B_t^{outside}$ of the outer ring. The image is cropped to this ROI, producing $I_t^{crop}$, which is fed into the segmentation model $f_{seg}$ to predict a binary mask $\hat{M}_t$ of break-up regions within the normalized coordinate system. The final segmentation map $\hat{S}_t$ in the original image coordinates is obtained by applying the inverse mapping function $\text{MapBack}$. Frames classified as 'Closed' are excluded from spatial analysis. This cascaded design ensures efficient and clinically interpretable TFBU quantification. 

The TF-Collab pipeline offers three advantages: temporal-spatial integration for joint BUT calculation and TFBU localization; computational efficiency via ROI cropping to boost segmentation performance; and direct clinical alignment that mirrors diagnostic workflows to enhance interpretability.

\subsection{TFBU Segmentation Problem and TF-Net}
\textbf{Problem Formulation and Challenges.}
The core segmentation task within the pipeline is defined on a single image. Let the input image space be $\mathcal{I} \subset \mathbb{R}^{H \times W \times 3}$ and the output space be $\mathcal{S} \subset \{0, 1\}^{H \times W}$, where 1 indicates the TFBU region. We can model the ideal segmentation mask $S$ as:
\begin{equation}
S(p) = \begin{cases}
1 & \text{if pixel } p \text{ exhibits break-up} \\
0 & \text{otherwise}
\end{cases}
\end{equation}
The segmentation model's goal is to learn a mapping:
\begin{equation}
f_{seg}: \mathcal{I} \rightarrow \mathcal{S}
\end{equation}

This task is particularly challenging due to several intrinsic difficulties: 1)\textbf{Extreme Class Imbalance}: As quantified in Section 3, TFBU pixels are extremely rare (e.g., $\sim$0.18\% in original images), making the model prone to predicting background. 2)\textbf{Amorphous and Diffuse Boundaries}: TFBU areas lack consistent shape, size, or texture. Their boundaries are often faint and irregular, defying simple geometric priors. The boundary complexity can be characterized by high perimeter-to-area ratios:

\begin{equation}
\mathcal{C}(S) = \frac{\text{Perimeter}(S)}{\text{Area}(S)} \gg 1
\end{equation}

3)\textbf{Multi-scale Characteristics}: Break-up areas appear at various scales, from microscopic spots to larger regions. This requires capturing features at multiple receptive fields simultaneously.

The complexity of this pixel-level decision rule necessitates a model capable of capturing rich, multi-scale contextual information to distinguish subtle pathological patterns from complex backgrounds.

\textbf{TF-Net Architecture.}
To address above the challenges, we propose \textbf{TF-Net}, as illustrated in Fig. \ref{fig:tf_dino}, a specialized architecture designed with three core components: an efficient feature extractor, multi-scale context aggregation, and progressive refinement. Let the input image be $I_0$. The TF-Net architecture can be formally defined as a composition of functions: 

\begin{equation}
\mathcal{F}_{TF\text{-}Net}(I_0) = \mathcal{F}_{dec} \circ \mathcal{F}_{ppm} \circ \mathcal{F}_{enc}(I_0)
\end{equation}

where $\mathcal{F}{enc}$ denotes the encoder, $\mathcal{F}{ppm}$ the pyramid pooling module, and $\mathcal{F}_{dec}$ the decoder.

\begin{figure*}[!t]
    \centering
    \includegraphics[width=1\textwidth]{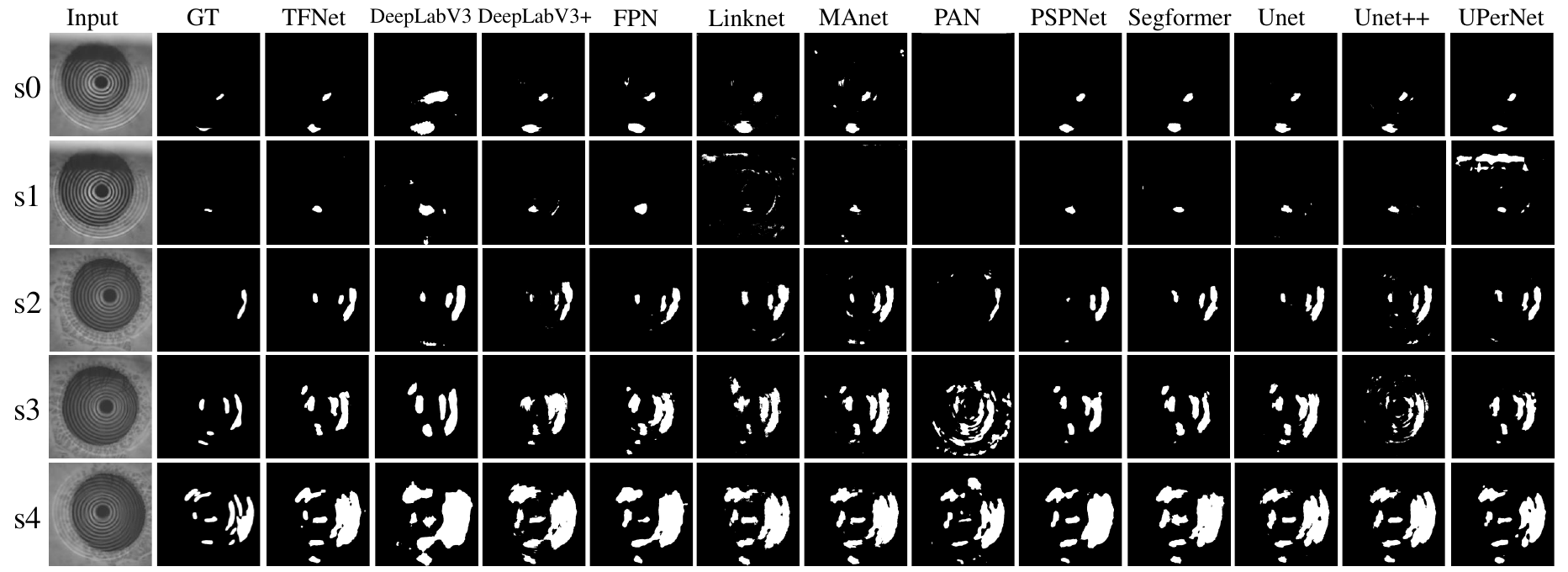}
    \caption{
    Visual comparison of segmentation results on the TF-Crop test set across five scale variants (s0-s4) of MobileOne\cite{mobileone} or MobileOne-mini(only for TF-Net). The proposed TF-Net demonstrates superior segmentation accuracy with more precise boundary delineation and fewer false positives and negatives compared to other baseline models.
    }
    \label{fig:Compare_result}
\end{figure*}

\textbf{Efficient Encoder with Structural Reparameterization}: 
The encoder $\mathcal{F}_{enc}$ employs MobileOne blocks with reparameterization technique. For a given layer with input $X$ and output $Y$, during training we have:
\begin{equation}
Y = \sum_{k=1}^{K} W_k \ast X + b_k
\end{equation}
where $K$ parallel branches learn diverse features. During inference, these are reparameterized into a single efficient operation:
\begin{equation}
Y = W_{fusion} \ast X + b_{fusion}
\end{equation}
This design provides training-time richness with inference-time efficiency, crucial for handling the complex TFBU patterns while maintaining clinical practicality.

\textbf{Multi-scale Context Aggregation}: 
To address the multi-scale nature of TFBU regions, we employ a Pyramid Pooling Module (PPM), similar to PSPNet\cite{pspnet}, that captures context at multiple scales:
\begin{equation}
P = \mathcal{F}_{ppm}(F_{enc}) = \text{Concat}\left[\text{AdaptivePool}_{s_i}(F_{enc})\right]_{i=1}^{4}
\end{equation}
where $s_i \in \{1,2,3,6\}$ represents different pooling scales. This ensures that both small and large break-up regions receive adequate contextual information.

\textbf{Progressive Refinement Decoder}: 
The decoder $\mathcal{F}_{dec}$ progressively recovers spatial details through skip connections:
\begin{equation}
F_{dec}^{l} = \mathcal{U}(F_{dec}^{l+1}) \oplus F_{enc}^{l}
\end{equation}
where $\mathcal{U}$ denotes upsampling, $\oplus$ represents feature fusion, and $l$ indicates feature level. This hierarchical refinement is essential for accurately capturing the irregular boundaries of TFBU regions.

The TF-Net architecture specifically addresses the TFBU segmentation challenges: For \textbf{class imbalance}, we incorporate class-frequency-aware optimization through weighted loss functions. Let $N_c$ be the number of pixels belonging to class $c \in \{0,1\}$ (background and TFBU), and $N_{total} = \sum_c N_c$ be the total number of pixels. The class weight for each category is calculated as:
\begin{equation}
w_c = \frac{N_{total}}{C \cdot N_c}
\end{equation}
where $C=2$ is the number of classes. These weights are normalized to ensure stable training:
\begin{equation}
\hat{w}_c = \frac{w_c}{\sum_{c} w_c}
\end{equation}
The weighted cross-entropy loss\cite{celoss} function then becomes:
\begin{equation}
\mathcal{L}_{WCE} = -\sum_{c=1}^{C} \hat{w}_c \cdot \frac{1}{N_c} \sum_{i=1}^{N_c} \log(p_{c}^{(i)})
\end{equation}
where $p_{c}^{(i)}$ is the predicted probability for the true class of pixel $i$. This formulation effectively counteracts the bias toward the dominant background class by assigning higher weights to the rare TFBU pixels.

For \textbf{irregular morphology}, the encoder-decoder structure with skip connections preserves spatial details necessary for boundary-aware segmentation:
\begin{equation}
\nabla\mathcal{F}_{TF\text{-}Net} \approx \nabla I \cdot \nabla S
\end{equation}
where the gradient flow maintains sensitivity to boundary variations.

In conclusion, TF-Net directly translates raw tear film video data into quantifiable, clinically actionable insights for dry eye diagnosis. By overcoming the specific challenges of TFBU segmentation, it provides the reliable foundation for the TF-Collab pipeline to automate the measurement of key biomarkers like break-up time and area.

For \textbf{multi-scale characteristics}, the multi-scale pyramid pooling provides enhanced multi-scale feature representation by aggregating contextual information across different receptive fields. The pyramidal feature extraction ensures that break-up regions of all sizes are effectively captured:
\begin{equation}
\mathcal{R}_{effective} = \bigcup_{i=1}^{n} \mathcal{R}(s_i)
\end{equation}
where $\mathcal{R}(s_i)$ denotes the receptive field at scale $s_i$.

\begin{table*}
  \centering
  \fontsize{8.5pt}{9.0pt}\selectfont
  \setlength{\heavyrulewidth}{1.5pt}
  \begin{tabular}{rccccccccc}
    \toprule
    Model & Backbone & Param(M) & Flops(G) & FPS & mIoU & mDSC & mRecall & HD95$\downarrow$ & ASSD$\downarrow$ \\
    & & train/infer & train/infer & GPU/CPU & & & & & \\
    \midrule
    Unet & mobileone s0 & 8.6/5.4 & 18.9/14.4 & 76.1/4.6 & 0.652 & 0.731 & 0.973 & 400.3 & 133.6 \\
    Unet++ & mobileone s0 & 9.9/6.7 & 45.4/41.0 & 32.0/2.4 & 0.610 & 0.684 & 0.969 & 579.3 & 209.8 \\
    Deeplabv3 & mobileone s0 & 12.8/9.6 & 52.1/38.3 & 54.6/2.8 & 0.472 & 0.527 & 0.933 & 548.3 & 257.7 \\
    Deeplabv3+ & mobileone s0 & 6.1/2.9 & 9.7/4.6 & 168.3/5.5 & 0.584 & 0.651 & 0.952 & 474.2 & 173.6 \\
    FPN & mobileone s0 & 6.3/3.1 & 14.1/9.7 & 124.3/5.1 & 0.597 & 0.669 & 0.949 & 374.2 & 132.3 \\
    PAN & mobileone s0 & 4.5/1.3 & 7.2/2.2 & 207.2/7.9 & 0.494 & 0.498 & 0.501 & 1001.3 & 798.1 \\
    PSPNet & mobileone s0 & 4.4/1.2 & 3.4/1.3 & 319.8/14.9 & 0.654 & 0.735 & 0.973 & 294.5 & 82.3 \\
    MANet & mobileone s0 & 36.2/33.0 & 24.7/20.2 & 59.7/4.0 & 0.581 & 0.649 & 0.958 & 631.0 & 243.6 \\
    LinkNet & mobileone s0 & 5.8/2.6 & 8.7/4.3 & 131.6/6.3 & 0.584 & 0.653 & 0.972 & 486.3 & 190.5 \\
    UPerNet & mobileone s0 & 14.3/11.1 & 58.9/54.5 & 44.3/2.8 & 0.630 & 0.705 & 0.956 & 417.6 & 144.1 \\
    Segformer & mobileone s0 & 5.0/1.8 & 12.6/8.1 & 57.6/4.5 & 0.672 & 0.752 & 0.969 & 344.7 & 102.5 \\
    \cellcolor{lightgray}TF-Net (Ours) & \cellcolor{lightgray}mobileone-mini0 & \cellcolor{lightgray}0.72/0.65 & \cellcolor{lightgray}2.7/2.4 & \cellcolor{lightgray}302.2/33.8& \cellcolor{lightgray}0.698 & \cellcolor{lightgray}0.779 & \cellcolor{lightgray}0.975 & \cellcolor{lightgray}212.9 & \cellcolor{lightgray}54.1 \\
    \midrule
    Unet & mobileone s1 & 9.2/9.1 & 19.9/19.5 & 61.7/3.5 & 0.645 & 0.725 & 0.977 & 413.8 & 129.7 \\
    Unet++ & mobileone s1 & 12.3/12.3 & 77.5/77.2 & 21.9/1.7 & 0.636 & 0.714 & 0.978 & 432.6 & 141.7 \\
    Deeplabv3 & mobileone s1 & 14.0/13.9 & 56.6/55.9 & 36.6/2.1 & 0.537 & 0.596 & 0.959 & 549.3 & 220.2 \\
    Deeplabv3+ & mobileone s1 & 5.7/5.6 & 8.4/8.1 & 115.1/4.4 & 0.582 & 0.650 & 0.958 & 471.3 & 185.6 \\
    FPN & mobileone s1 & 5.7/5.6 & 13.2/12.9 & 96.0/4.2 & 0.612 & 0.687 & 0.968 & 350.1 & 118.3 \\
    PAN & mobileone s1 & 3.9/3.9 & 6.1/5.7 & 130.7/5.7 & 0.535 & 0.583 & 0.799 & 496.5 & 217.3 \\
    PSPNet & mobileone s1 & 3.8/3.7 & 2.7/2.4 & 222.2/11.8 & 0.668 & 0.747 & 0.915 & 328.6 & 88.9 \\
    MANet & mobileone s1 & 53.3/53.2 & 31.0/30.7 & 43.2/3.1 & 0.626 & 0.702 & 0.973 & 478.0 & 167.6 \\
    LinkNet & mobileone s1 & 6.2/6.1 & 10.7/10.4 & 87.4/4.7 & 0.560 & 0.617 & 0.816 & 612.1 & 268.1 \\
    UPerNet & mobileone s1 & 14.5/14.5 & 58.2/57.8 & 40.0/2.7 & 0.649 & 0.725 & 0.953 & 404.8 & 136.5 \\
    Segformer & mobileone s1 & 4.4/4.4 & 11.9/11.6 & 50.8/3.9 & 0.665 & 0.746 & 0.963 & 275.5 & 79.1 \\
    \cellcolor{lightgray}TF-Net (Ours) & \cellcolor{lightgray}mobileone-mini1 & \cellcolor{lightgray}2.6/2.3 & \cellcolor{lightgray}8.9/7.9 & \cellcolor{lightgray}166.5/17.5 & \cellcolor{lightgray}0.711 & \cellcolor{lightgray}0.792 & \cellcolor{lightgray}0.976 & \cellcolor{lightgray}182.0 & \cellcolor{lightgray}47.8 \\
    \midrule
    Unet & mobileone s2 & 13.6/13.5 & 24.8/24.4 & 53.4/3.2 & 0.655 & 0.735 & 0.975 & 351.9 & 106.9 \\
    Unet++ & mobileone s2 & 18.1/18.0 & 89.1/88.7 & 19.6/1.5 & 0.621 & 0.697 & 0.974 & 461.0 & 163.4 \\
    Deeplabv3 & mobileone s2 & 22.0/21.9 & 88.6/87.8 & 25.6/1.5 & 0.591 & 0.662 & 0.975 & 387.3 & 130.0 \\
    Deeplabv3+ & mobileone s2 & 9.0/8.9 & 12.3/11.9 & 87.6/3.9 & 0.616 & 0.691 & 0.961 & 396.8 & 129.8 \\
    FPN & mobileone s2 & 8.2/8.2 & 15.9/15.5 & 80.7/3.8 & 0.625 & 0.700 & 0.950 & 329.2 & 103.7 \\
    PAN & mobileone s2 & 6.4/6.3 & 9.2/8.8 & 95.3/5.2 & 0.525 & 0.549 & 0.562 & 471.1 & 228.8 \\
    PSPNet & mobileone s2 & 6.2/6.1 & 3.8/3.6 & 181.2/10.1 & 0.666 & 0.748 & 0.966 & 292.7 & 79.1 \\
    MANet & mobileone s2 & 126.1/126.1 & 52.6/52.2 & 29.1/2.3 & 0.636 & 0.712 & 0.967 & 441.6 & 151.9 \\
    LinkNet & mobileone s2 & 12.1/12.0 & 17.4/17.0 & 68.6/4.2 & 0.624 & 0.701 & 0.973 & 428.5 & 149.4 \\
    UPerNet & mobileone s2 & 19.4/19.3 & 61.2/60.9 & 36.4/2.5 & 0.676 & 0.753 & 0.954 & 337.1 & 104.0 \\
    Segformer & mobileone s2 & 7.0/6.9 & 14.6/14.2 & 46.2/3.6 & 0.668 & 0.749 & 0.967 & 304.1 & 81.3 \\
    \cellcolor{lightgray}TF-Net (Ours) & \cellcolor{lightgray}mobileone-mini2 & \cellcolor{lightgray}5.7/5.2 & \cellcolor{lightgray}19.8/17.6 & \cellcolor{lightgray}83.2/9.4 & \cellcolor{lightgray}0.720 & \cellcolor{lightgray}0.800 & \cellcolor{lightgray}0.973 & \cellcolor{lightgray}180.8 & \cellcolor{lightgray}47.0 \\
    \midrule
    Unet & mobileone s3 & 16.3/16.2 & 28.9/28.5 & 46.7/2.9 & 0.656 & 0.737 & 0.977 & 397.0 & 113.4 \\
    Unet++ & mobileone s3 & 22.9/22.8 & 113.6/113.2 & 16.6/1.3 & 0.631 & 0.702 & 0.834 & 549.9 & 190.6 \\
    Deeplabv3 & mobileone s3 & 24.3/24.2 & 98.1/97.2 & 22.6/1.4 & 0.611 & 0.686 & 0.974 & 366.4 & 113.5 \\
    Deeplabv3+ & mobileone s3 & 11.3/11.2 & 15.7/15.2 & 71.7/3.5 & 0.647 & 0.726 & 0.961 & 292.5 & 84.9 \\
    FPN & mobileone s3 & 10.6/10.5 & 19.4/18.9 & 68.0/3.6 & 0.636 & 0.714 & 0.962 & 305.5 & 96.0 \\
    PAN & mobileone s3 & 8.7/8.6 & 12.9/12.4 & 77.5/4.5 & 0.521 & 0.570 & 0.808 & 613.0 & 314.5 \\
    PSPNet & mobileone s3 & 8.6/8.5 & 5.4/5.1 & 147.0/8.4 & 0.706 & 0.786 & 0.952 & 234.7 & 60.4 \\
    MANet & mobileone s3 & 129.5/129.4 & 57.7/57.3 & 26.5/1.0 & 0.661 & 0.741 & 0.978 & 367.4 & 108.6 \\
    LinkNet & mobileone s3 & 14.7/14.6 & 22.8/22.3 & 56.9/3.6 & 0.636 & 0.714 & 0.973 & 378.6 & 123.6 \\
    UPerNet & mobileone s3 & 21.8/21.7 & 64.7/64.2 & 33.5/2.4 & 0.657 & 0.734 & 0.952 & 360.0 & 123.3 \\
    Segformer & mobileone s3 & 9.3/9.2 & 18.0/17.6 & 41.7/3.1 & 0.684 & 0.765 & 0.970 & 242.7 & 70.3 \\
    \cellcolor{lightgray}TF-Net (Ours) & \cellcolor{lightgray}mobileone-mini3 & \cellcolor{lightgray}10.2/9.2 & \cellcolor{lightgray}35.1/31.3 & \cellcolor{lightgray}64.3/4.2 & \cellcolor{lightgray}0.721 & \cellcolor{lightgray}0.801 & \cellcolor{lightgray}0.972 & \cellcolor{lightgray}187.5 & \cellcolor{lightgray}51.9 \\
    \midrule
    Unet & mobileone s4 & 21.5/21.4 & 36.2/35.6 & 32.5/2.4 & 0.664 & 0.744 & 0.975 & 361.1 & 101.7 \\
    Unet++ & mobileone s4 & 33.4/33.3 & 171.4/170.8 & 11.4/0.9 & 0.673 & 0.753 & 0.977 & 331.6 & 92.7 \\
    Deeplabv3 & mobileone s4 & 29.0/28.9 & 111.4/110.2 & 12.9/1.1 & 0.583 & 0.653 & 0.967 & 420.5 & 143.2 \\
    Deeplabv3+ & mobileone s4 & 16.1/16.0 & 21.8/21.1 & 41.9/2.8 & 0.629 & 0.705 & 0.963 & 320.2 & 108.1 \\
    FPN & mobileone s4 & 15.4/15.3 & 25.6/25.0 & 42.4/3.2 & 0.644 & 0.724 & 0.964 & 285.7 & 83.6 \\
    PAN & mobileone s4 & 13.6/13.5 & 19.4/18.8 & 43.0/3.3 & 0.527 & 0.566 & 0.785 & 710.7 & 395.1 \\
    PSPNet & mobileone s4 & 13.6/13.5 & 9.6/9.1 & 106.6/6.4 & 0.664 & 0.746 & 0.965 & 310.7 & 83.4 \\
    MANet & mobileone s4 & 135.7/135.5 & 66.7/66.1 & 21.0/1.8 & 0.629 & 0.704 & 0.975 & 423.7 & 153.0 \\
    LinkNet & mobileone s4 & 20.0/19.9 & 33.1/32.5 & 36.0/2.9 & 0.647 & 0.727 & 0.979 & 364.0 & 120.7 \\
    UPerNet & mobileone s4 & 26.6/26.5 & 70.9/70.3 & 25.9/2.1 & 0.642 & 0.720 & 0.942 & 368.3 & 130.2 \\
    Segformer & mobileone s4 & 14.2/14.1 & 24.3/23.7 & 30.4/2.3 & 0.670 & 0.750 & 0.973 & 286.4 & 83.3 \\
    \cellcolor{lightgray}TF-Net (Ours) & \cellcolor{lightgray}mobileone-mini4 & \cellcolor{lightgray}15.9/14.4 & \cellcolor{lightgray}54.6/48.8 & \cellcolor{lightgray}38.3/3.0 & \cellcolor{lightgray}0.737 & \cellcolor{lightgray}0.814 & \cellcolor{lightgray}0.965 & \cellcolor{lightgray}175.6 & \cellcolor{lightgray}44.8 \\
    \bottomrule
  \end{tabular}
    \caption{
    TFBU segmentation performance comparison of various models on the TF-Crop test set. Our TF-Net utilizes MobileOne-mini backbone - a further streamlined version of the original MobileOne\cite{mobileone} architecture where stage-wise block repetitions are reduced from [2, 8, 10, 1] to [2, 3, 4, 3]. Additionally, we strategically constrain the channel dimensions across network layers to further minimize parameter count and computational complexity. These architectural refinements collectively enable substantially accelerated inference speeds on mobile platforms while maintaining competitive segmentation accuracy.
    }
  \label{tab:seg_main}
\end{table*}
\section{Experiment}
\subsection{Experiments Setup}

\begin{table*}
  \centering
  \fontsize{8.0pt}{8.5pt}\selectfont
  \setlength{\heavyrulewidth}{1.5pt}
  \begin{tabular}{cccccccccccc}
    \toprule
     Resolution & Param(M) & Flops(G) & Crop & TFBU\_IoU & mIoU & TFBU\_DSC & mDSC & TFBU\_Recall & mRecall & HD95$\downarrow$ & ASSD$\downarrow$ \\
    \midrule
     \multirow{2}{*}{512*512} & \multirow{2}{*}{0.651} & \multirow{2}{*}{2.383}
    & $\times$ & 0.066 & 0.528 & 0.119 & 0.557 & 0.461 & 0.726 & 557.3 & 271.7 \\
    &&& \checkmark & 0.407 & 0.698 & 0.564 & 0.779 & 0.961 & 0.975 & 212.9 & 54.1 \\
    \midrule
     \multirow{2}{*}{1024*1024} & \multirow{2}{*}{0.651} & \multirow{2}{*}{9.529}
    & $\times$ & 0.266 & 0.631 & 0.406 & 0.702 & 0.943 & 0.970 & 337.9 & 112.8 \\
    &&& \checkmark & 0.423 & 0.711 & 0.581 & 0.799 & 0.972 & 0.985 & 186.6 & 49.7 \\
    \bottomrule
  \end{tabular}
  \caption{
    Ablation study on the effect of cropping strategy and input resolution on segmentation performance.
}
  \label{tab:ablation_crop}
\end{table*}

\begin{table*}
  \centering
  \fontsize{8.0pt}{8.5pt}\selectfont
  \setlength{\heavyrulewidth}{1.5pt}
  \begin{tabular}{lccccccccc}
    \toprule
    Model & Weighted & TFBU\_IoU & mIoU & TFBU\_DSC & mDSC & TFBU\_Recall & mRecall & HD95$\downarrow$ & ASSD$\downarrow$ \\
    \midrule
    TF-Net & $\times$ & 0.081 & 0.536 & 0.135 & 0.565 & 0.100 & 0.5497 & 5952 & 279.9    \\
    TF-Net & \checkmark & 0.407 & 0.698 & 0.564 & 0.779 & 0.961 & 0.975 & 212.9 & 54.1 \\
    \bottomrule
  \end{tabular}
  \caption{
    Ablation study comparing segmentation performance with and without class-weighted loss.
}
  \label{tab:ablation_weighted}
\end{table*}

\textbf{Implementation Details.}
The model processes input images at $512\times512$ resolution and generates segmentation masks of corresponding spatial dimensions. All models are trained for 50 epochs using the SGD optimizer with a momentum of 0.937, an initial learning rate of $1\times10^{-2}$, and weight decay of $5\times10^{-4}$. We implement a ReduceLROnPlateau scheduler that reduces the learning rate by a factor of 0.5 when the validation loss fails to improve for 3 consecutive epochs.

The class weighting scheme effectively addresses the extreme class imbalance (99.07\% background vs. 0.93\% TFBU pixels). Class weights $\mathbf{w} = [w_0, w_1]$ are computed during training initialization as $w_c = \frac{N_{\text{total}}}{C \cdot N_c}$
where $N_{\text{total}}$ is the total number of pixels in the training set, $C=2$ is the number of classes, and $N_c$ is the number of pixels belonging to class $c$. These weights are then normalized and applied to the CrossEntropy loss\cite{celoss} function, ensuring stable training and improved segmentation performance for the minority TFBU class.

\textbf{Evaluation Metrics.}
To comprehensively evaluate the segmentation performance of our proposed method, we employ five established metrics that assess different aspects of segmentation quality. For volumetric overlap and spatial similarity, we utilize the Dice Similarity Coefficient (DSC) and Intersection over Union (IoU), which measure the agreement between predicted and ground truth regions. Boundary accuracy is quantified using the95th Percentile Hausdorff Distance (HD95) and Average Symmetric Surface Distance (ASSD), evaluating the distance between segmentation contours. Clinical relevance is further assessed through Recall, measuring the model's ability to correctly identify tear film break-up regions. This multi-faceted evaluation protocol provides a holistic assessment of segmentation performance from both technical and clinical perspectives.

\subsection{Comparative Experiments}
We conducted extensive comparative experiments, as shown in \ref{fig:Compare_result}, against numerous state-of-the-art semantic segmentation architectures. The selected baselines encompass both classical and recently proposed models, spanning a diverse range of design paradigms, including U-Net \cite{unet}, FPN \cite{fpn}, PSPNet \cite{pspnet}, LinkNet \cite{linknet}, MAnet \cite{manet}, UNet++ \cite{unet++}, DeepLabV3 \cite{deeplabv3}, DeepLabV3+ \cite{deeplabv3plus}, PAN \cite{pan}, UPerNet \cite{upernet}, and SegFormer \cite{segformer}. To align with the requirements for practical mobile deployment, we adopted MobileOne \cite{mobileone}—a lightweight network incorporating re-parameterization techniques and tailored for mobile devices—as the backbone for all models. All models were trained on the training set of TF-Crop, with the TF-Crop validation set used to monitor the training process. Final performance evaluation was conducted on the held-out TF-Crop test set. This comprehensive model selection enables a thorough evaluation across different architectural philosophies and facilitates a clear assessment of the advantages of our proposed method in tear film segmentation. The main results are summarized in Table \ref{tab:seg_main}.

As presented in Table \ref{tab:seg_main}, alongside standard medical segmentation metrics, we specifically compared model parameters, FLOPs, and FPS(frames per second) on both a GPU (RTX 3050 Ti) and a CPU (Intel Core i5). This facilitates a holistic assessment of both segmentation accuracy and practical deployment potential. The results clearly demonstrate that our proposed TF-Net not only surpasses all other models across various segmentation metrics but also achieves significantly faster inference speeds, attaining a processing rate of 318.2(on GPU)/33.8(on CPU) frames per second. This level of performance adequately fulfills the requirements for real-time clinical diagnosis.

\subsection{Ablation Study}
To systematically validate the design choices in our proposed framework, we conduct comprehensive ablation studies examining four critical aspects: (1) the impact of the cropping strategy on addressing class imbalance; (2) the effectiveness of class-weighted loss functions; (3) the contribution of the Pyramid Pooling Module (PPM) for multi-scale feature extraction; and (4) the importance of the encoder-decoder architecture with skip connections for boundary refinement. All ablation experiments are performed using TF-Net with MobileOne-mini s0 as the backbone to ensure consistent and fair comparisons.
\subsubsection{Impact of Cropping Strategy for Class Imbalance Mitigation}
The severe class imbalance between Tear Film Break-Up (TFBU) regions (0.18\%) and background (99.82\%) in the original full-resolution images (3632 × 2760) poses a fundamental challenge for segmentation. Our proposed cropping strategy, which utilizes bounding boxes from the "Outside" detection stage to generate the TF-Crop subset, effectively alleviates this imbalance by increasing the relative proportion of TFBU pixels and directing model attention to the clinically relevant area.

As summarized in Table \ref{tab:ablation_crop}, this strategy leads to substantial gains across all segmentation metrics. The improvement is particularly notable at higher input resolutions, where cropping significantly enhances both the detection of TFBU regions and the accuracy of boundary delineation. A consistent positive trend is also observed at lower resolutions. The resolution-dependent nature of these gains highlights that combining high-resolution input with targeted cropping is essential for capturing the subtle textural characteristics of TFBUs. These findings confirm cropping as a crucial preprocessing step for robust tear film segmentation.

\subsubsection{Effectiveness of Class-Weighted Loss Function}
To address the severe class imbalance at the optimization level, we employ a class-weighted cross-entropy loss that assigns higher weights to the minority TFBU category. As quantitatively demonstrated in Table \ref{tab:ablation_weighted}, the proposed loss function yields dramatic improvements across all critical segmentation metrics.

For our TF-Net, the class-weighted loss brings remarkable performance gains confirming that the class-weighted loss effectively counteracts the model's bias toward the dominant background class, enabling balanced learning between TFBU and non-TFBU regions. The strategy proves to be a crucial complement to our spatial cropping approach, jointly addressing class imbalance from both the input and optimization perspectives to achieve robust segmentation of rare TFBU patterns.

\subsubsection{Contribution of Pyramid Pooling Module}
The segmentation of TFBU regions presents two interconnected challenges: the multi-scale appearance of break-up areas, which range from tiny scattered spots to larger connected regions, and their amorphous, diffuse boundaries. To address these, TF-Net incorporates a PPM Module for multi-scale context aggregation and an encoder-decoder structure with skip connections for boundary refinement. Ablation studies validate the necessity of both components.

As shown in Tab.\ref{tab:ablation_ppm_skip}, Removing the PPM leads to a consistent decline for HD95 and ASSD, while other key metrics show only negligible decreases. Similarly, eliminating skip connections severely impacts boundary precision. This degradation highlights the importance of skip connections in preserving spatial details for accurately delineating irregular TFBU contours. The encoder-decoder architecture with skip connections maintains effective gradient flow for boundary learning ($\nabla\mathcal{F}_{TF\text{-}Net} \approx \nabla I \cdot \nabla S$), which is crucial for clinical assessment of tear film stability.

\begin{table*}
  \centering
  \fontsize{8.0pt}{8.5pt}\selectfont
  \setlength{\heavyrulewidth}{1.5pt}
  \begin{tabular}{lcccccccccccc}
    \toprule
    Model & PPM & Skip & TFBU\_IoU & mIoU & TFBU\_DSC & mDSC & TFBU\_Recall & mRecall & HD95$\downarrow$ & ASSD$\downarrow$ \\
    \midrule
    TF-Net &  & & 0.388 & 0.688 & 0.546 & 0.770 & 0.952 & 0.970 & 235.2 & 59.8 \\
    TF-Net & \checkmark  & & 0.380 & 0.684 & 0.538 & 0.766 & 0.942 & 0.965 & 194.2 & 54.7 \\
    TF-Net & & \checkmark & 0.402 & 0.695 & 0.559 & 0.777 & 0.946 & 0.968 & 215.4 & 57.6 \\
    TF-Net & \checkmark & \checkmark & 0.407 & 0.698 & 0.564 & 0.779 & 0.961 & 0.975 & 212.9 & 54.1 \\
    \bottomrule
  \end{tabular}
  \caption{
    Ablation study evaluating the contributions of PPM module and Skip connections.
  }
  \label{tab:ablation_ppm_skip}
\end{table*}

\begin{table*}
  \centering
  \fontsize{8.0pt}{8.5pt}\selectfont
  \setlength{\heavyrulewidth}{1.5pt}
  \begin{tabular}{lccccccccccc}
    \toprule
    Testset & TFBU\_IoU & mIoU & TFBU\_DSC & mDSC & TFBU\_Recall & mRecall & TFBU\_FPR & mFPR & HD95$\downarrow$ & ASSD$\downarrow$ \\
    \midrule
    Broken-only & 0.407 & 0.698 & 0.564 & 0.779 & 0.961 & 0.975 & 0.011 	& 0.025 & 212.9 & 54.1 \\
    Non-Broken & 0.000 & 0.488 & 0.000 & 0.494 & 0.000 & 0.488 & 0.024 & 0.012 & N/A & N/A \\
    Full-Test & 0.068 & 0.523 & 0.094 & 0.541 & 0.159 & 0.569 & 0.022 & 0.014  & 212.9 & 54.1 \\
    \bottomrule
  \end{tabular}
  \caption{
  Robustness evaluation of TF-Net across different test subsets of TF-Crop:  Broken-only set, Non-Broken set and Full-Test set.
  }
  \label{tab:robustness_eval}
\end{table*}

\subsection{System Robustness and TF-Collab Evaluation}
The deployment of an AI-assisted diagnostic system in a clinical setting demands not only high accuracy under ideal conditions but also robust reliability when confronted with real-world variability and potential upstream errors. To this end, we conducted a comprehensive evaluation that assesses both the inherent robustness of the TF-Net segmentation model and the practical efficacy of the integrated TF-Collab pipeline. This analysis specifically targets a critical failure mode: the propagation of errors from the frame classification module to the segmentation stage.

\textbf{Inherent Robustness of the Segmentation Model.} We first designed a stress test to evaluate TF-Net's behavior when presented with inputs that should ideally be filtered out by a perfect classifier. The model was evaluated on three distinct test sets: the standard Broken-only set, a Non-Broken set (containing Clear and Blur frame counterparts from the TF-Crop subset), and the complete Full-Test set. The results, summarized in Table \ref{tab:robustness_eval}, provide compelling insights.

TF-Net demonstrates exceptional robustness by exhibiting near-perfect specificity on the Non-Broken set. The TFBU-class IoU and Dice scores of 0.0 indicate that the model virtually never hallucinates break-up regions in frames without pathology. This is further corroborated by a very low TFBU false positive rate (FPR) of 0.024. This property is crucial as it acts as a "safety net" for the integrated system; even if the classification model $f_{cls}$ errs by passing a non-pathological frame to the segmenter, TF-Net will not produce a misleading segmentation output, thereby preventing a cascade towards misdiagnosis.

Conversely, on the Broken-only set, TF-Net achieves a high TFBU recall of 0.961, confirming its strong sensitivity in detecting genuine break-up events. The performance on the Full-Test set (mDSC: 0.541, mIoU: 0.523) represents a realistic balance between its high performance on broken frames and the conservative behavior on non-broken frames, accurately reflecting its expected behavior in a clinical video stream.

\textbf{End-to-End Performance of the TF-Collab Pipeline.} Building upon the robust foundation of TF-Net, we evaluated the complete TF-Collab pipeline on full-length, unseen clinical videos. The pipeline seamlessly integrated the frame classification, placido rings detection, and TFBU segmentation models to fully automate the diagnostic workflow.

The pipeline successfully processed input videos in real-time, achieving an average throughput of 318.2(on GPU, 3050ti)/33.8(on CPU, core i5) frames per second. Its end-to-end segmentation performance for frames identified as Broken yielded a mDSC of 0.779, which is only marginally lower than TF-Net's standalone performance on the curated Broken-only test set. This minor performance drop can be attributed to occasional inaccuracies in the placido rings detection, leading to sub-optimal cropping, rather than failures of the segmentation core itself.

\section{Conclusion}
\label{sec:conclusion}
This work addresses the critical need for automated tear film analysis by introducing the Tear Film Multi-task (TFM) Dataset, the first comprehensive benchmark for automated tear film analysis, and proposing TF-Net, an efficient segmentation model for tear film break-up analysis. We further design TF-Collab, an integrated pipeline that synergistically combines multiple vision tasks to enable objective and reproducible diagnostic assessment.

While the current approach employs separate models for classification, detection, and segmentation, future work will explore unified multi-task learning architectures. A single model trained jointly on all three tasks could potentially improve performance through shared representations while reducing system complexity. Additional directions include expanding the dataset scale and diversity, and validating the system in clinical trials.

\bibliographystyle{IEEEtran}
\bibliography{references.bib}
\end{document}